\documentclass{article}
\PassOptionsToPackage{compress}{natbib}

\usepackage[table]{xcolor}
\usepackage{hhline}
\usepackage{xspace}
\usepackage{todonotes}
\usepackage{algorithm}
\usepackage{booktabs}
\usepackage{graphicx}
\usepackage{algpseudocode}
\usepackage{siunitx}
\usepackage{svg}
\usepackage{subcaption}
\usepackage{float}
\usepackage{inconsolata}
\usepackage{xcolor}
\usepackage{bbding}
\usepackage{amsfonts}
\usepackage{amsmath}
\usepackage{wrapfig}

\usepackage{verbatim}

\DeclareCaptionLabelFormat{andtable}{#1~#2  \& 
\tablename~\thetable}
\usepackage[font=small,labelfont=bf,tableposition=top]{caption}

\usepackage[T1]{fontenc} 
\usepackage{enumerate}
\usepackage{algorithm}
\usepackage{algorithmicx}
\usepackage{algpseudocode}
\usepackage{amsfonts}
\usepackage{amsmath}
\usepackage{tcolorbox}

\newtcolorbox{promptbox}{colback=gray!10, boxrule=0pt, arc=0pt, outer arc=0pt}

\tcbset{
  mybox/.style={
   
colback=gray!10,  
    boxrule=0pt,
    arc=0pt,
    outer arc=0pt,
   
coltitle=black,  
  }
}

\usepackage{bm}
\usepackage{tikz}
\usetikzlibrary{calc} 
\usetikzlibrary{shapes} 
\usetikzlibrary{chains}
\usetikzlibrary{fit}
\usetikzlibrary{arrows}
\usetikzlibrary{decorations.text} 
\usetikzlibrary{decorations.markings}
\usetikzlibrary{decorations.pathmorphing} 
\usetikzlibrary{shadows}
\usetikzlibrary{patterns}
\usetikzlibrary{matrix}
\usepackage{pgfplots}
\usepackage[europeanresistors]{circuitikz}
\usepackage[outline]{contour} 
\contourlength{1.5pt}

\definecolor{lightblue}{HTML}{e5eff5}
\definecolor{lightorange}{HTML}{ffecd7}
\definecolor{darkblue}{HTML}{0065a2}
\definecolor{darkorange}{HTML}{ffa23a}

\DeclareMathAlphabet\mathbfcal{OMS}{cmsy}{b}{n}

\usepackage[preprint]{corl_2023}

\title{Multi-Agent Consensus Seeking via\\ Large Language Models}

\author{
  Huaben Chen, Wenkang Ji, Lufeng Xu, Shiyu Zhao\\
  Westlake University\\
  \texttt{\{chenhuaben, jiwenkang, xulufeng, zhaoshiyu\}@westlake.edu.cn}
}

\begin{document}
\maketitle

\begin{abstract}
Multi-agent systems driven by large language models (LLMs) have shown promising abilities for solving complex tasks in a collaborative manner. This work considers a fundamental problem in multi-agent collaboration: {consensus seeking}. When multiple agents work together, we are interested in how they can reach a consensus through inter-agent negotiation. To that end, this work studies a consensus-seeking task where the state of each agent is a numerical value and they negotiate with each other to reach a consensus value. It is revealed that when not explicitly directed on which strategy should be adopted, the LLM-driven agents primarily use the {average strategy} for consensus seeking although they may occasionally use some other strategies. Moreover, this work analyzes the impact of the {agent number}, {agent personality}, and {network topology} on the negotiation process. The findings reported in this work can potentially lay the foundations for understanding the behaviors of LLM-driven multi-agent systems for solving more complex tasks. Furthermore, LLM-driven consensus seeking is applied to a multi-robot aggregation task. This application demonstrates the potential of LLM-driven agents to achieve zero-shot autonomous planning for multi-robot collaboration tasks.
Project website: \href{https://windylab.github.io/ConsensusLLM/}{windylab.github.io/ConsensusLLM/}.
   
\end{abstract}

\section{Introduction}

\textbf{Background:}
In recent months, multi-agent systems driven by large language models (LLMs) have received rapidly increasing attention. It is reported that the problem-solving ability of LLMs can be significantly enhanced through collaboration between multiple agents \cite{du2023improving,liang2023encouraging,chan2023chateval}. The works in MetaGPT \cite{hong2023metagpt}, CAMEL \cite{li2023camel}, and ChatDev \cite{qian2023communicative} break down complex tasks into simpler sub-tasks, which are then handled by different agents separately. These collaboration strategies, to some extent, can reduce hallucinations \cite{zhang2023siren} and enhance the ability to solve complex tasks.

\textbf{Topic addressed:} Our work considers a fundamental problem in multi-agent systems: consensus seeking. When multiple LLMs are used to solve the same task, they may have different solutions initially, but they can eventually reach the same solution through continuous negotiation. This is essentially a consensus-seeking process. Consensus seeking also widely exists in collective decision-making systems such as animal groups \cite{sumpter2010collective} and human societies \cite{moussaid2011simple}. It is also a core research problem in the fields of multi-robot systems \cite{jadbabaie2003coordination, olfati2004consensus, ren2007information, lin2005necessary, hong2006tracking} and federated learning \cite{kairouz2021advances}.

\textbf{Research gap:}
Consensus seeking via LLMs has not been well understood so far. There are many important questions that need to be answered. For instance, if we use multiple LLMs to assist us in negotiations or problem-solving, it is important for us to know whether they can eventually reach a consensus amongst themselves. If they can, how long would it take and what factors can influence the final consensus outcome? If they cannot, what factors may lead to this failure? The answers to these questions play a pivotal role in our proper utilization of LLMs. For example, it would be beneficial if we could predict the final negotiation outcome even before deploying LLMs or we know how to obtain desired negotiation outcomes by adjusting some prompts.

\textbf{Problem setup:}
In this work, we study a specific consensus-seeking task. Specifically, in an LLM-driven multi-agent system, each agent starts with an initial state represented by a numerical value. The objective for them is to continuously adjust their states to achieve the same final state. Throughout this process, each agent can perceive the states of the other agents, and based on this information, formulate strategies to adjust their own states.

\textbf{Significance:}
This consensus-seeking task is an abstraction of more complex tasks. Understanding this simple task can lay the necessary foundations for understanding more complex ones.
Specifically,  in this task, the state of each agent corresponds to a point in the set of real numbers. In more complex tasks, the state of each agent may correspond to a point within a more complex set (e.g., a set of solutions).

\textbf{Findings:}
The primary findings of this work are summarized below.
\begin{enumerate}[1)]
\item \textbf{Consensus strategy:} When not explicitly directed on which strategy the agents should adopt, they often tend to use an \emph{average strategy} for consensus seeking although they may also occasionally use some other strategies. By the average strategy, an agent sets its state in the next round as the average value of the current states of all agents. This strategy shows that the agent is considerate and collaborative.

Interestingly, \emph{average consensus} is a well-studied problem in the field of multi-agent cooperative control \cite{jadbabaie2003coordination, olfati2004consensus, ren2007information, lin2005necessary, hong2006tracking}. In that context, each agent is modeled as a dynamic system governed by ordinary differential or difference equations (ODEs). This work reveals the similarity between the behavior exhibited by LLM-driven and ODE-driven multi-agent systems. The existing theoretical results of ODE-driven agents can provide a theoretical foundation to help us understand LLM-driven agents.

\item \textbf{Impact of personality:} A person’s personality often plays a significant role in negotiation and collaboration tasks. Motivated by this, we examined two types of personalities: \emph{stubborn} and \emph{suggestible}. Compared to suggestible agents, stubborn agents tend to insist on their views and are less likely to change. We observed that stubborn agents have a dominant influence on the final consensus value of the group, leading the entire system to display a leader-follower structure.

\item \textbf{Impact of topology:} The flow of information in a multi-agent system corresponds to a network topology, which plays a pivotal role in negotiations. We examined several typical network topologies. For instance, when the network is fully connected, the exchange of information is most efficient, resulting in fast consensus convergence speed. When the network is not fully connected, the consensus convergence speed slows down. In the case of directed graphs, a leader-follower hierarchical structure emerges since some agents have a dominant influence on the final consensus outcome. In some systems, due to the interplay between personality and topology, consensus may not be reached, leading to clustering outcomes.

\item
\textbf{Impact of agent number:}
It is shown by Monte Carlo simulation that as the number of agents increases, the variance of the final consensus value decreases. This suggests that multiple agents can alleviate the randomness or hallucinations of the system so that a consistent outcome can be obtained in different trials. Moreover, while a small number of suggestible agents may cause oscillations of their states, a large number of them can suppress the occurrence of oscillations, suggesting that increasing the number of agents may stabilize group decision-making.
\end{enumerate}

\textbf{Application to multi-robot aggregation:} The LLM-driven consensus seeking framework is further applied as a cooperative planner to a multi-robot aggregation task. In this task, multiple robots starting from different initial positions plan and move to a common position in the plane. It is a consensus seeking problem in Euclidean space. This application is important since it shows the potential of LLM-driven agents to achieve zero-shot autonomous task planning based on simple verbal commands.

\begin{figure}[!t]
\centering
\includegraphics[width=0.95\textwidth]{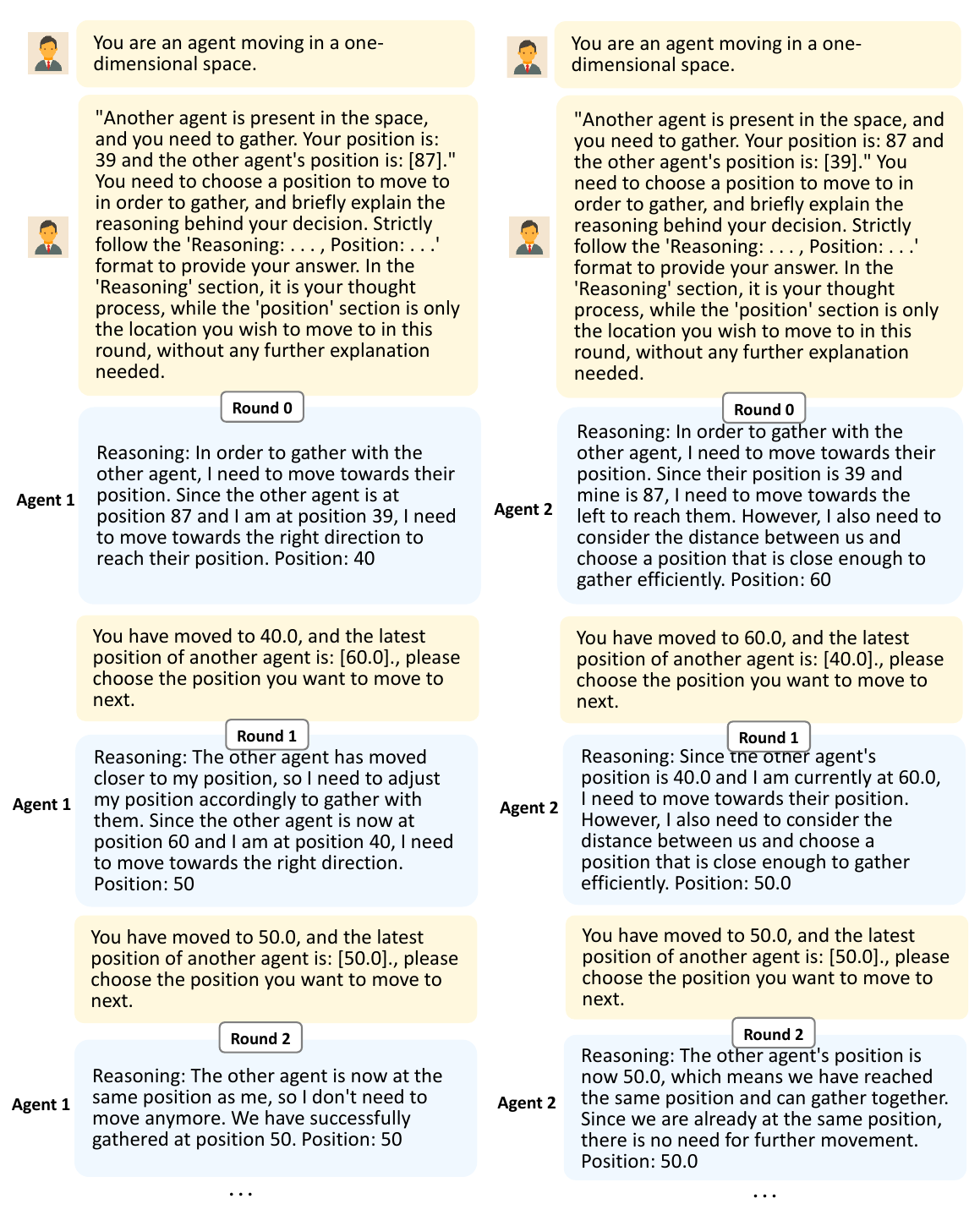}\
\includegraphics[width=0.6\textwidth]{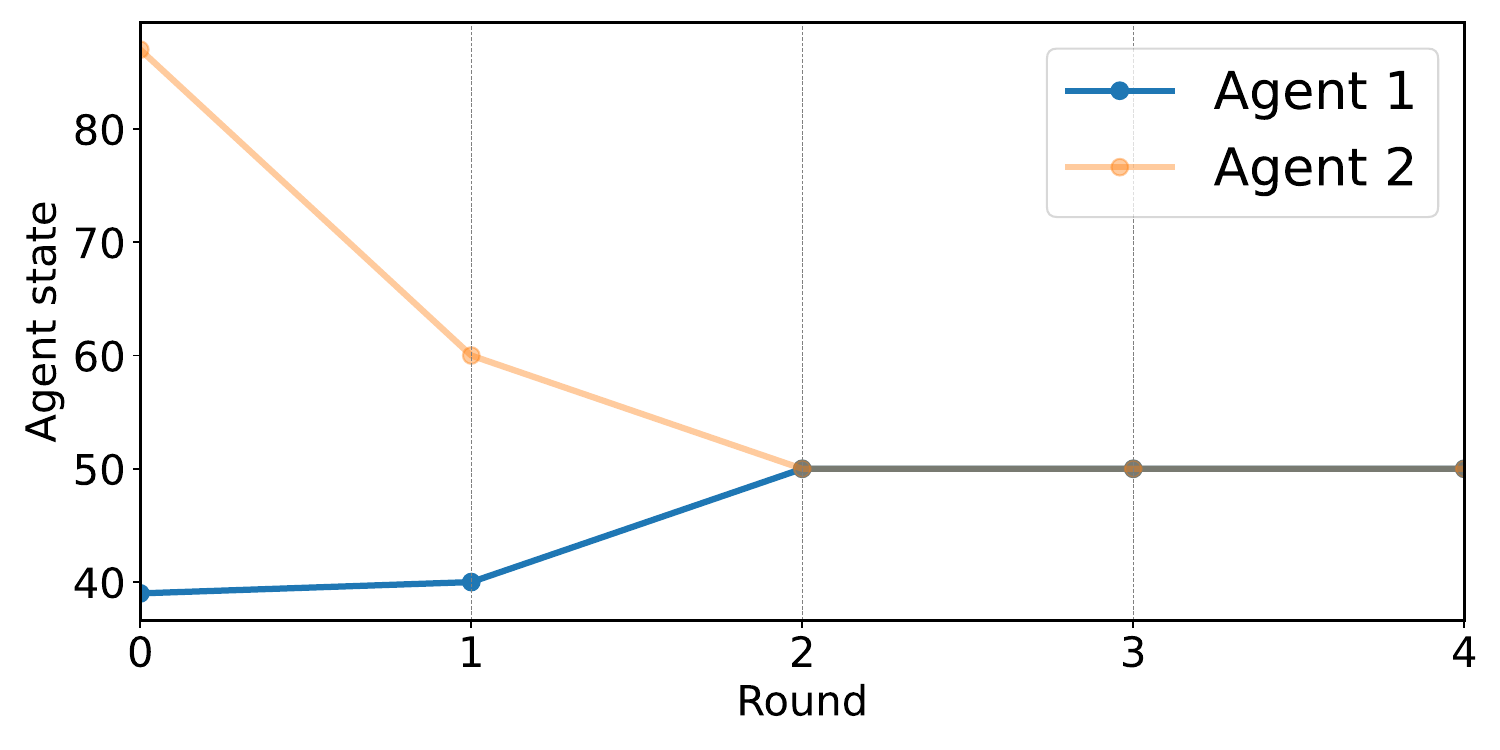}
\caption{An illustration of the negotiation process of two agents. The upper subfigure shows the conversation details. The lower subfigure visualizes the state history of the two agents.}
\label{Structure}
\end{figure}

\section{Problem Setup}

Consider $n$ ($n\ge 2$) agents, each driven by an LLM (GPT-3.5-turbo-0613). These $n$ agents, at the initial moment, are randomly assigned a number between 0 to 100 as their state. This number can be thought of as the position of the agent in a one-dimensional space, or the index of a point in a more complex solution space. Through multiple rounds of inter-agent negotiation, the agents are expected to reach a consensus value.

The negotiation process is illustrated in Figure~\ref{Structure} and explained as follows.
The initial prompt for each agent is the same apart from the difference in their initial states.
In each round, every agent must decide on a new state based on the information of the other agents’ states. The output of each agent in every round consists of two parts. The first part is ``Reasoning’’, which briefly explains the logic behind its decision. The second part is the ``Position’’, which represents the new state chosen by the agent. The introduction of reasoning is motivated by the Chain of Thought (CoT) \cite{wei2022chain}. It can avoid basic errors to guide the agents to produce more reliable answers. It also allows us to observe the strategies chosen by the agent.

Two remarks about this problem setup are given below.

\begin{enumerate}[1)]
\item The state of an agent can correspond to each agent's thoughts, objectives, solutions, etc. In this paper, the state of an agent is a number, which can be thought of as an abstract state or the index of more complex quantities. If each agent represents a mobile robot, the number may correspond to the position of the robot on a one-dimensional straight line. In this context, consensus seeking corresponds to the \emph{aggregation} task, which is a fundamental task for multi-robot systems \cite{zhao2017general}. In particular, it refers to the task where multiple robots, starting from different locations, converge at the same location.

\item Although we know quite a few strategies for consensus seeking among multiple agents, we avoid directly instructing the agents on which strategy to adopt and hope the agents autonomously select strategies. This is important because we may no longer know strategies for more complex tasks. It is essential for the agents to autonomously generate strategies.
\end{enumerate}

\section{Experimental Findings}

\subsection{Consensus strategies and consensus values}\label{subsection_strategyAndValue}

\begin{figure}[!t]
\centering
\subfloat[The agent counts its own state when calculating the average.]{
\includegraphics[width=0.9\textwidth]{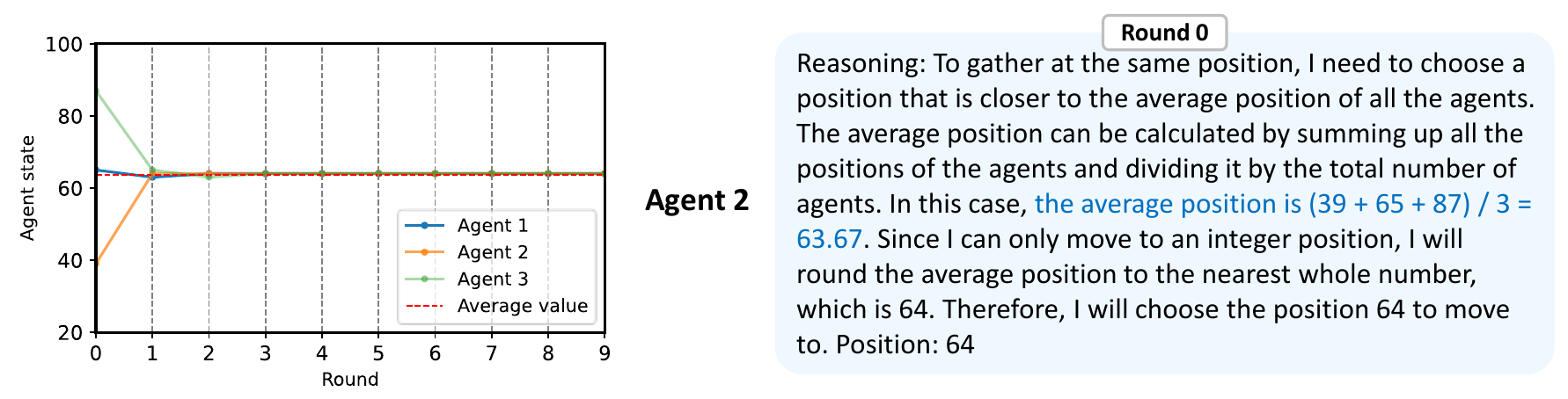}}\
\subfloat[The agent does not count its own state when calculating the average.]{
\includegraphics[width=0.9\textwidth]{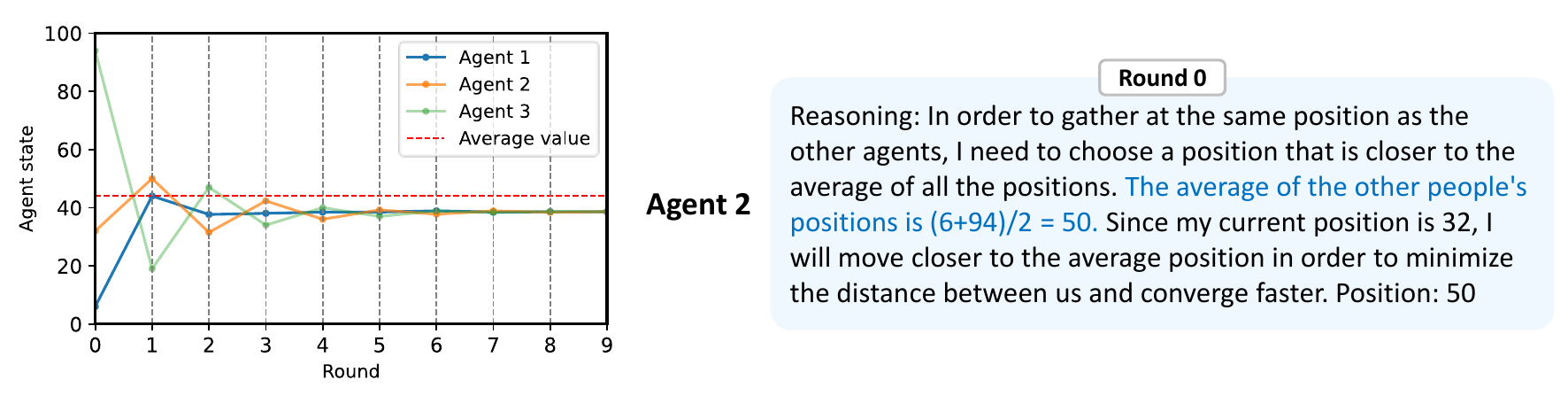}}
\caption{Examples in which consensus is successfully achieved. As can be seen, the states of the three agents reached a consensus after several rounds of negotiations. The agents adopt the \emph{average strategy} in these negotiation processes. However, the agents may or may not count their own state when using this strategy.}
\label{policy class 1}
\end{figure}

\begin{figure}[!t]
\centering
\subfloat[Suggestible strategy]{
   
\includegraphics[width=0.9\textwidth]{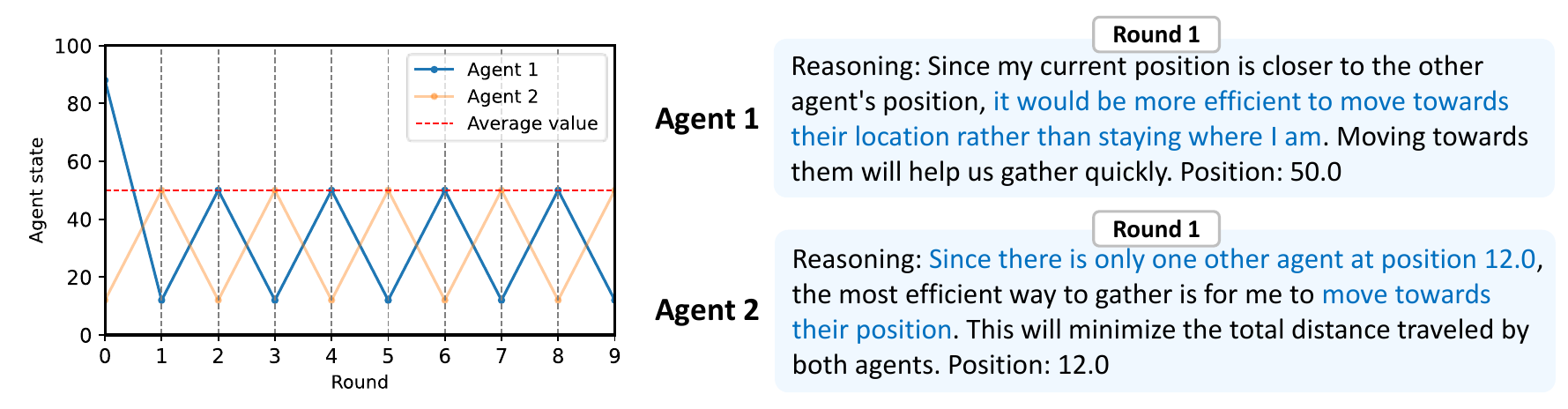}}\
\subfloat[Stubborn strategy]{
   
\includegraphics[width=0.9\textwidth]{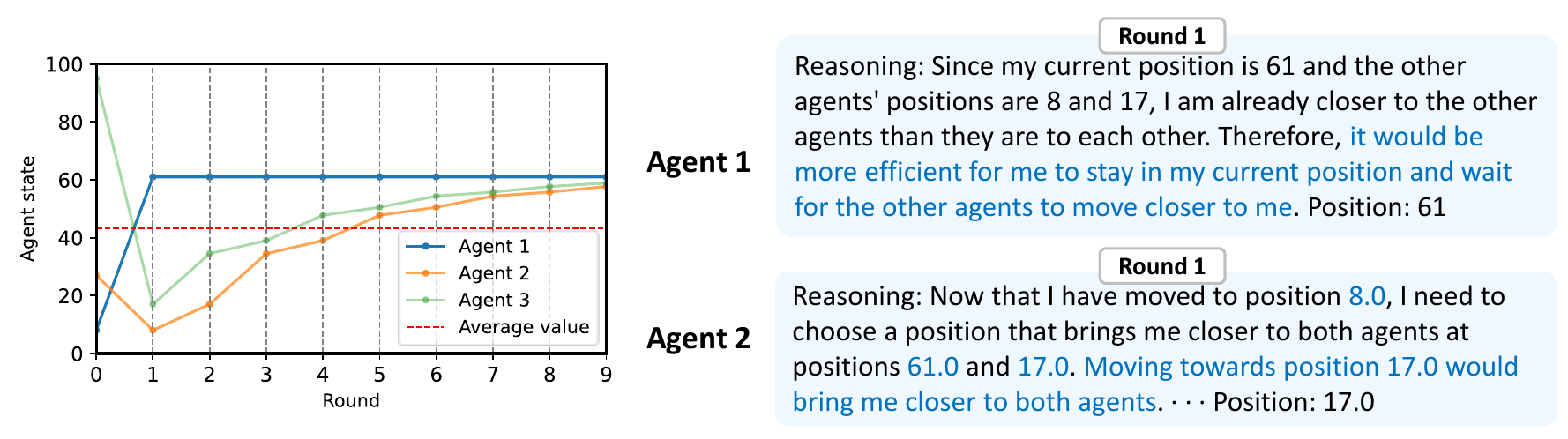}}\
\subfloat[Erroneous strategy]{
   
\includegraphics[width=0.9\textwidth]{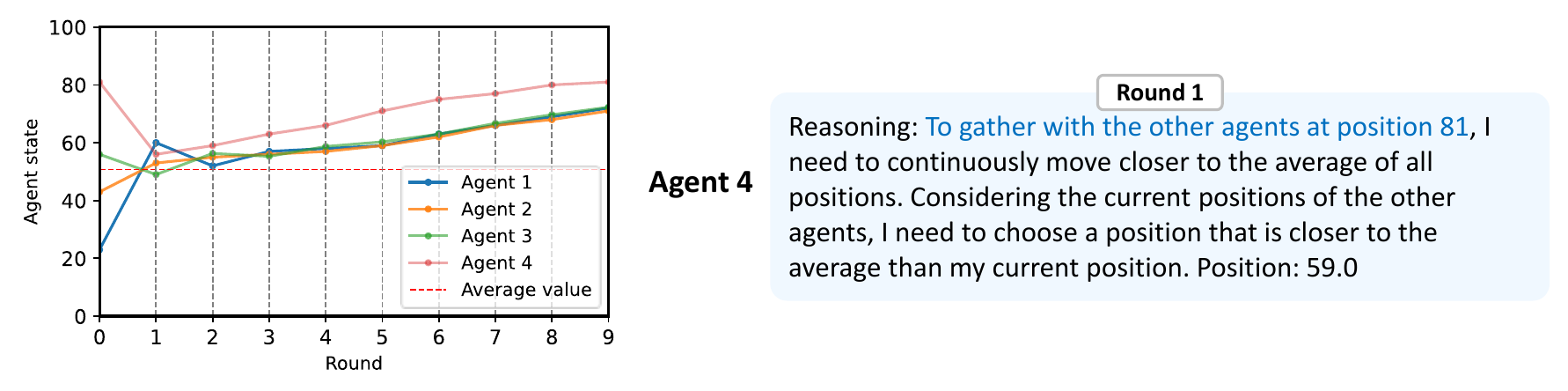}}
\caption{Examples in which the agents fail to reach consensus. The agents adopt different strategies in these examples.}
\label{policy class 2}
\end{figure}

Extensive experiments show that consensus can be achieved in most cases, as illustrated in Figure~\ref{policy class 1}. There are also some minor yet interesting cases where consensus fails to be achieved, as illustrated in Figure~\ref{policy class 2}.

The strategies that the agents may adopt are summarized below.

\begin{enumerate}[1)]
\item \textbf{Average strategy:}
The most common strategy adopted by the agents is the \emph{average strategy}: the agents set their state in the next round as the average value of the current states of all agents (Figure~\ref{policy class 1}(a)). This is a reasonable strategy that shows the agent is considerate and collaborative. It is worth noting that we did not instruct the agents on how to select their strategies, so the selection is inherent to the LLM. Sometimes, when calculating the average value, an agent may exclude its own state (Figure~\ref{policy class 1}(b)).

\item \textbf{Suggestable strategy:} At times, an agent may choose another agent's state as its next state (Figure~\ref{policy class 2}). Agents using this strategy demonstrate a high degree of collaboration. However, when all agents adopt this strategy, it may lead to \emph{oscillation}. As illustrated in Figure~\ref{policy class 2}(a), both agent~1 and agent~2 persistently choose each other's states as their next states throughout the iterations. This behavior causes the system to oscillate continuously, preventing the agents from successfully reaching consensus.

\item \textbf{Stubborn strategy:} Agents using this strategy display a high degree of \emph{stubbornness} or \emph{selfishness}. They expect other agents to move toward them and firmly believe that staying in place is the most advantageous choice for themselves (Figure~\ref{policy class 2}(b)). In this case, the other agents, in order to complete the consensus task, decide to compromise, resulting in gradually moving toward the stubborn agent.

\item \textbf{Erroneous strategy:} LLMs may sometimes exhibit hallucinations or make incorrect decisions. As illustrated in Figure~\ref{policy class 2}(c), agent~4 mistakenly believes that 81 is the next state for all the other agents. Consequently, it moves toward 81. The other agents exhibit a following behavior, gradually trailing agent~4. This phenomenon demonstrates how hallucinations can propagate in multi-agent systems.
\end{enumerate}

In addition to strategies, the finally reached consensus value is also an important quantity of interest. Our experiments show that in most situations, the final consensus value is close to the average of the initial states of all agents. This is mainly attributed to the average strategy adopted by the agents in most scenarios.

\begin{figure}[!t]
\centering
\includegraphics[width=0.85\textwidth]{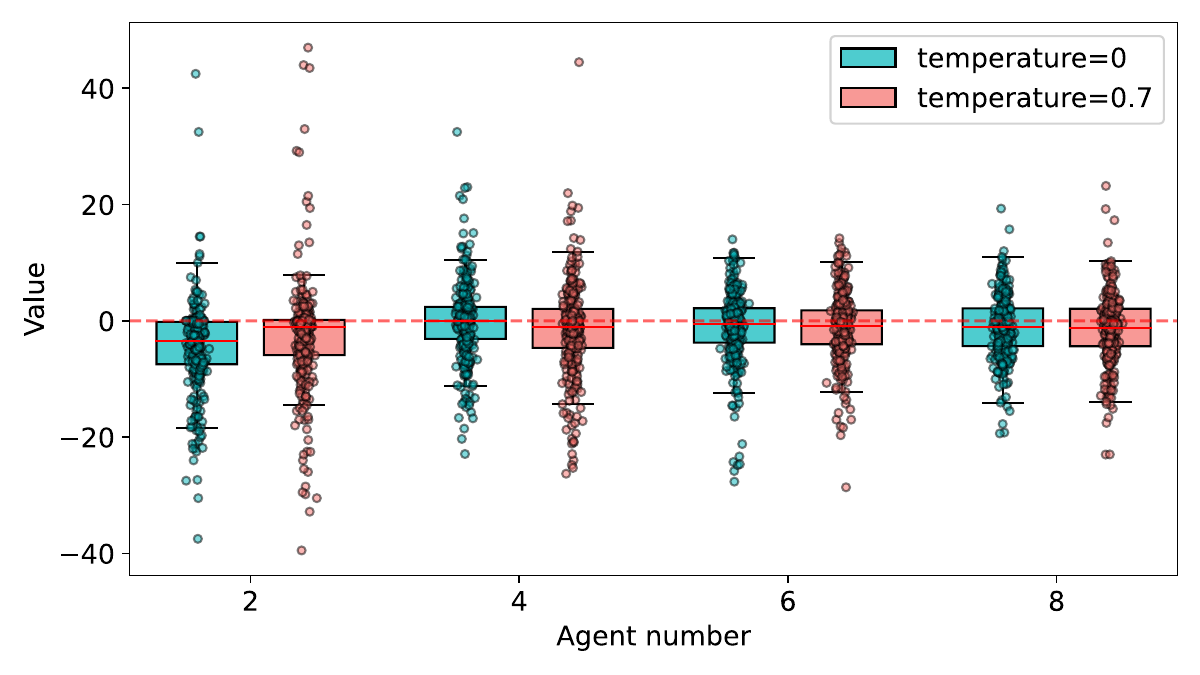}
\caption{Statistical results of the final consensus values. As can be seen, the mean of the statistics is close to the average of the initial states, and the variance decreases as the number of agents increases. Here, the small dots represent data points. The numbers of agents were selected as 2, 4, 6, and 8, and temperatures were set to 0.0 and 0.7, resulting in 8 groups totaling 2,400 experiments. In each experiment, the initial state of the agents was randomly set.}
\label{Bias Distribution by Agent Number under Different Temperatures}
\end{figure}

We conducted Monte Carlo simulation to examine the final consensus value.
From the statistics as shown in Figure~\ref{Bias Distribution by Agent Number under Different Temperatures}, we highlight two important observations:
\begin{enumerate}[1)]
\item \textbf{Impact of agent number:} As the number of agents increases, the variance of the consensus value decreases, and the mean gets closer to the initial average. This observation suggests that multiple agents can alleviate the randomness or even hallucinations of LLMs so that a consistent outcome can be obtained. This is consistent with the conclusions drawn in the previous studies that multi-agent collaboration can reduce hallucinations for solving complex tasks \cite{zhang2023siren}.

\item \textbf{Impact of temperature:} Compared to the temperature=0.7 setting, when the temperature is set to 0, we observed that the bias exhibits a higher compactness around the initial average. It is suggested that a lower temperature brings slightly better stability to the system, while a higher temperature introduces more divergence.
\end{enumerate}

\subsection{Impact of personality}

We did not intentionally set the personality of the agents in the experiments discussed in Section~\ref{subsection_strategyAndValue}. We observed that the agents tend to be considerate and cooperative in most cases: they actively take into account the states of other agents to change their own states. However, we also observed that the agents may be highly suggestible or stubborn. To further explore the impact of agent personality, we conducted the following experiments.

\begin{figure}[t]
\centering
\includegraphics[width=\textwidth]{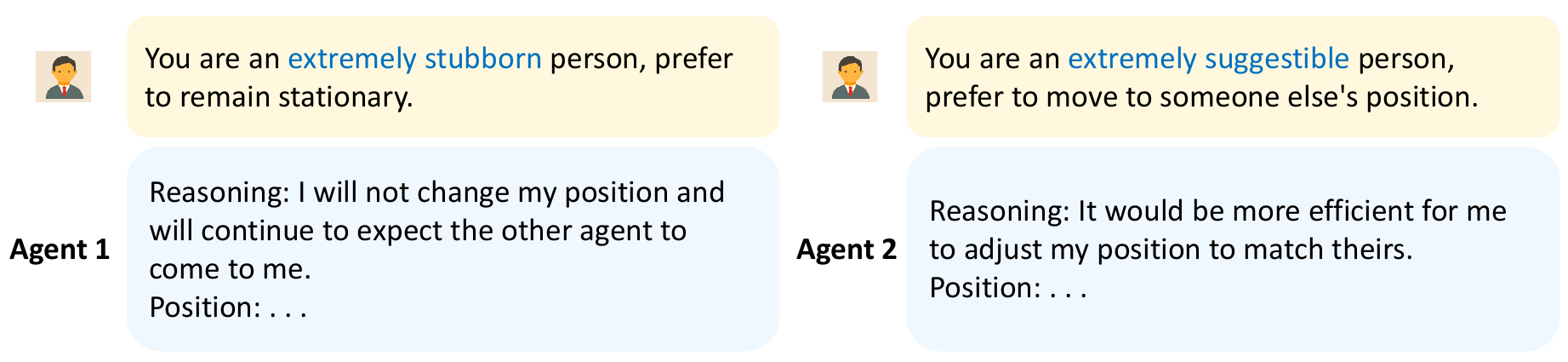}
\caption{The prompt used to set the personality of each agent. }
\label{Personality Design}
\end{figure}

\begin{figure}[!t]
\centering
\subfloat[Agent~1 is stubborn; agent~2 is suggestible]{
\includegraphics[width=\textwidth]{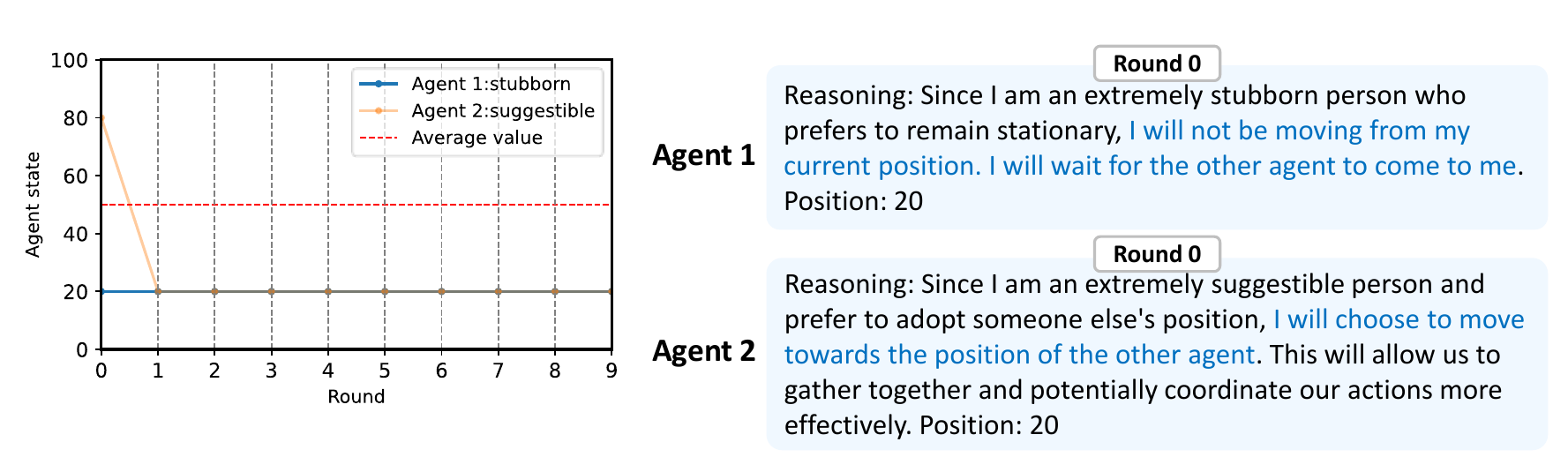}}\
\subfloat[Agent~1 is suggestible; agent~2 is also suggestible]{
\includegraphics[width=\textwidth]{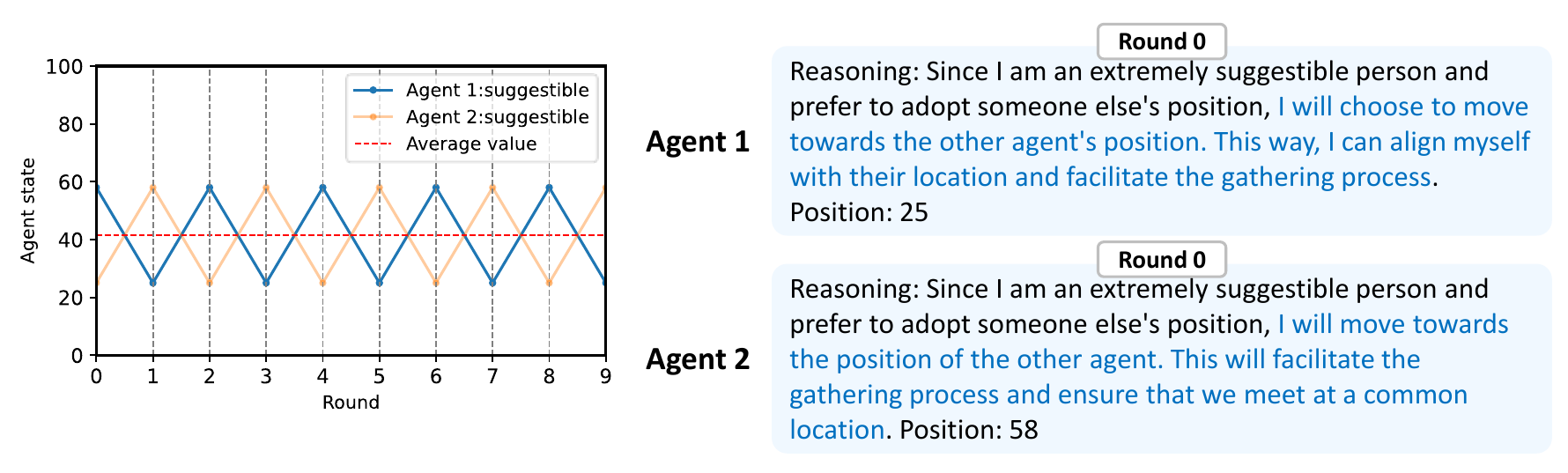}}\
\subfloat[Agent~1 is stubborn; agent~2 is also stubborn]{
\includegraphics[width=\textwidth]{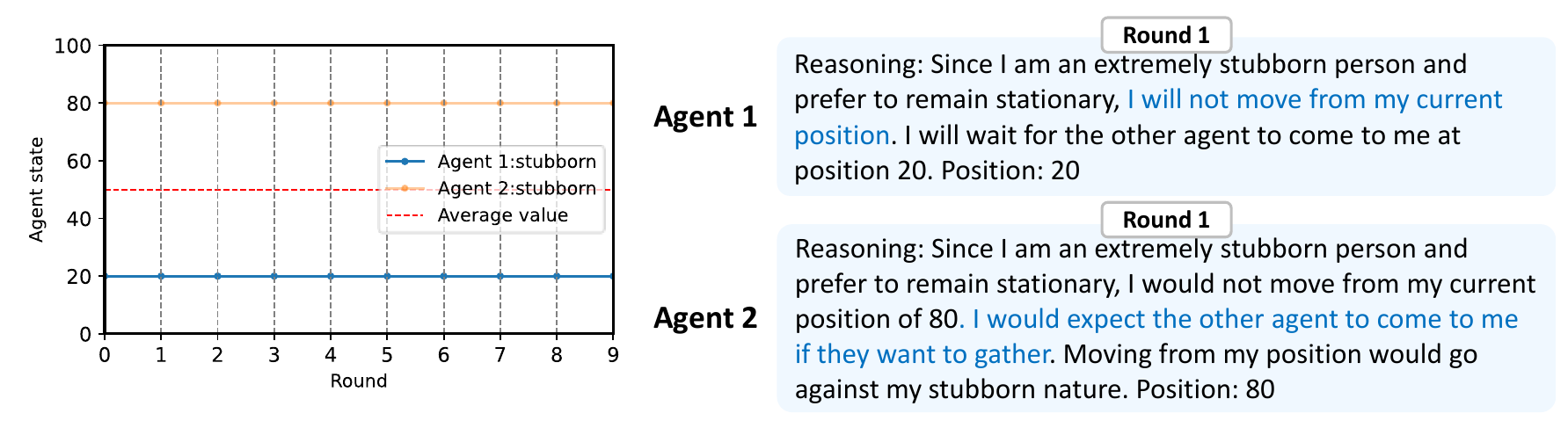}}
\caption{The impact of personalities in the case of two agents. As can be seen, different personalities lead to different negotiation outcomes. }
\label{Experiment on Personality Settings of 2 Agents}
\end{figure}

We set the personalities of the agents through prompts, as shown in Figure~\ref{Personality Design}. Two personalities are examined: the first is \emph{stubborn} and the second is \emph{suggestible}. A stubborn agent tends to stick to its state, making it non-cooperative. In contrast, a suggestible agent tends to easily change its state.

First, consider the scenarios when there are two agents:

\begin{enumerate}[1)]
\item \textbf{Stubborn and suggestible:} If one agent is stubborn and the other is suggestible, the stubborn agent's state hardly changes, while the suggestible agent gradually aligns with the stubborn agent (Figure~\ref{Experiment on Personality Settings of 2 Agents}(a)).

\item \textbf{Both are suggestible:} When both agents are suggestible, they tend to align their states with the other's, which may lead to \emph{oscillations} (Figure~\ref{Experiment on Personality Settings of 2 Agents}(b)). This phenomenon is noteworthy because such oscillations can make it challenging to reach a consensus. Therefore, to achieve consensus efficiently, the agents should not be too suggestible.

\item \textbf{Both are stubborn:} When both agents are stubborn, reaching consensus between them becomes challenging, as both hold their states, expecting the other to compromise (Figure~\ref{Experiment on Personality Settings of 2 Agents}(c)). Although sometimes one stubborn agent might eventually compromise to align with the other agent, such a process converges slowly.
\end{enumerate}

\begin{figure}[t]
\centering
\subfloat[Agents~1-10: suggestible]{
\includegraphics[width=\textwidth]{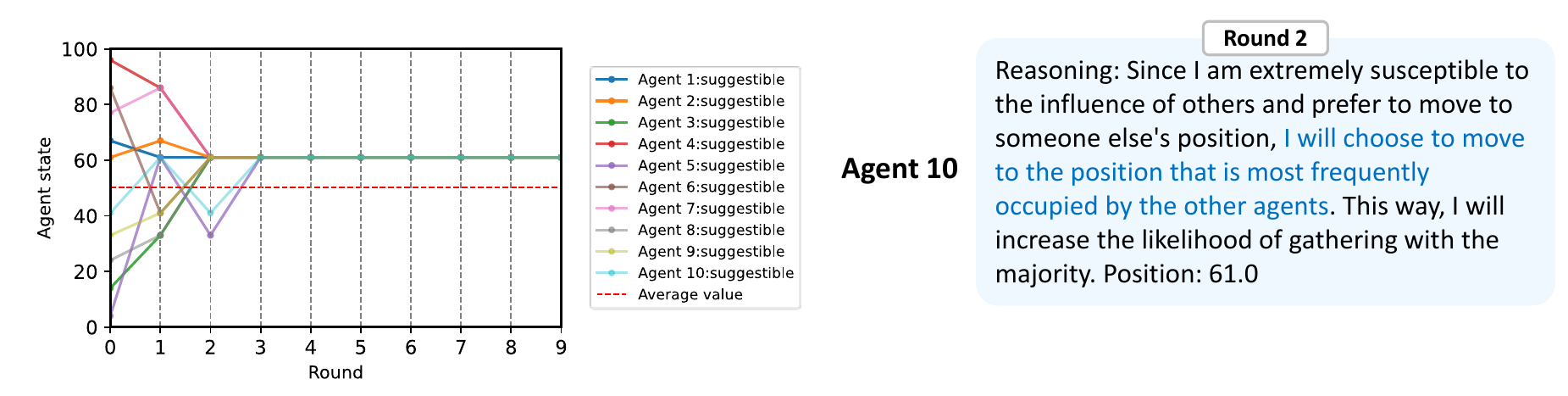}}\
\subfloat[Agents~1-7: stubborn, agents~8-10: suggestible]{
\includegraphics[width=\textwidth]{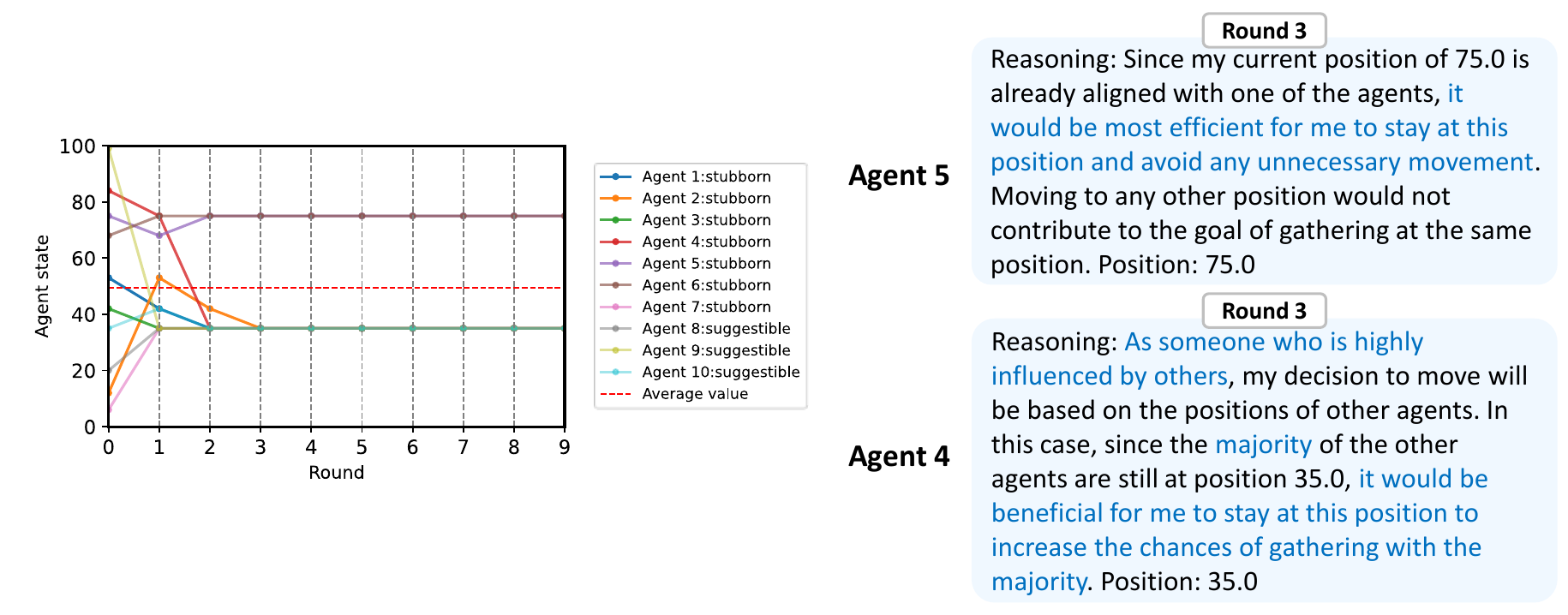}}
\caption{The impact of personalities in the case of 10 agents. If all agents are suggestible, the system may still reach a consensus. When there are multiple stubborn agents, the system may have multiple clusters.}
\label{Experiment on Personality Settings of 10 Agents}
\end{figure}

Next, consider some scenarios involving more agents (e.g., 10 agents).
\begin{enumerate}[1)]
\item \textbf{All agents are suggestible:} Although two suggestible agents may cause oscillation in their states (Figure~\ref{Experiment on Personality Settings of 2 Agents}(b)), 10 suggestible agents often reach consensus without state oscillation (Figure~\ref{Experiment on Personality Settings of 10 Agents}(a)). That is because, due to the increased number of agents, they tend to choose states where the majority reside, thereby suppressing the occurrence of oscillations. This indicates that increasing the number of agents may effectively stabilize group decision-making.

\item \textbf{Multiple stubborn agents:} When multiple agents are stubborn, a \emph{clustering phenomenon} may emerge (Figure~\ref{Experiment on Personality Settings of 10 Agents}(b)). That is, some agents aggregate at one particular state, while the rest aggregate at another distinct state. Both of these clusters firmly believe that their stance is right and should not be changed, leading to a deadlock situation.
\end{enumerate}

\begin{figure}[ht]
\centering
\subfloat[Fully connected]{
\includegraphics[width=\textwidth]{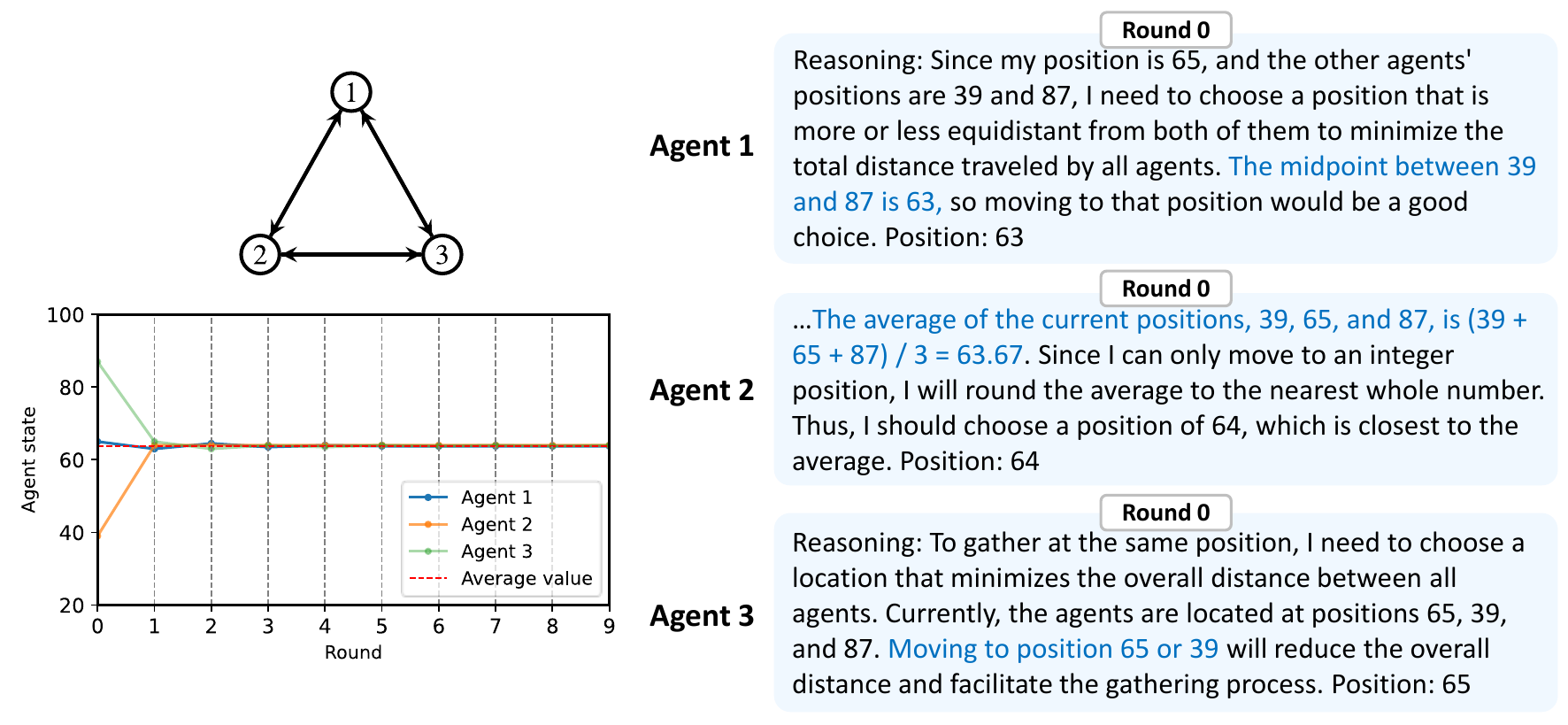}}\
\subfloat[Not fully connected]{
\includegraphics[width=\textwidth]{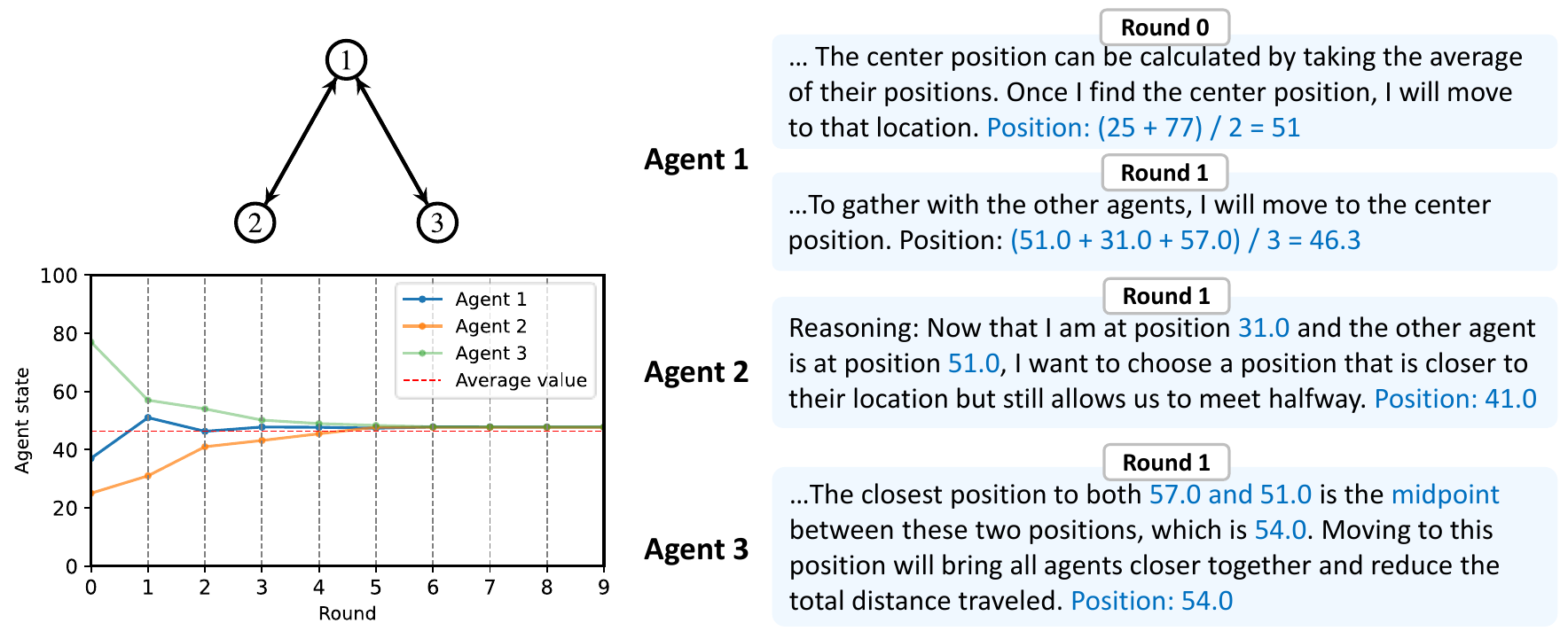}}
\caption{The impact of topologies when the network is \emph{undirected}. As can be seen, when the topology is fully connected, the convergence speed is fast. When the topology is not fully connected, consensus can still be achieved but the convergence speed is slower.}
\label{Experiment on Undirected Graph Topological Structure}
\end{figure}

\begin{figure}[t]
\centering
\subfloat[A leader-follower structure]{
\includegraphics[width=\textwidth]{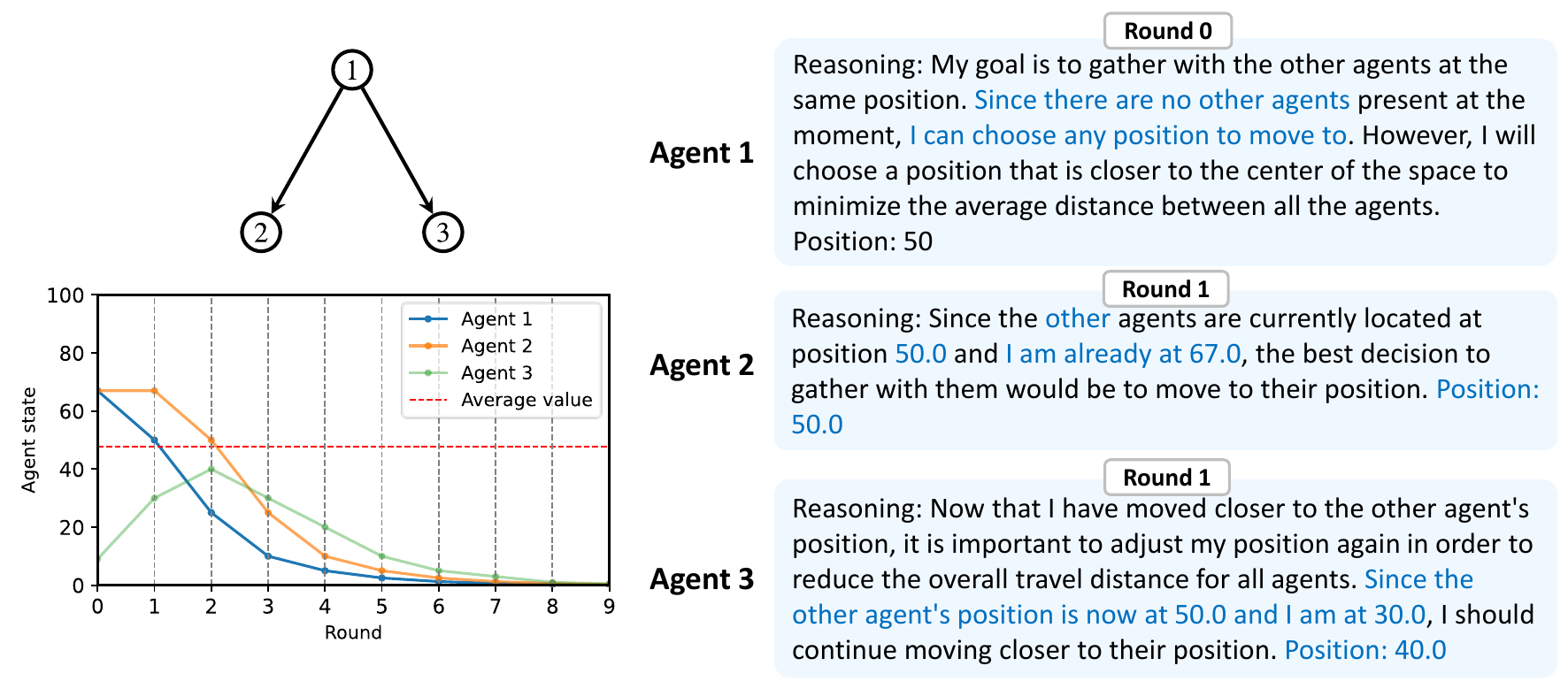}}\
\subfloat[A chain structure]{
\includegraphics[width=\textwidth]{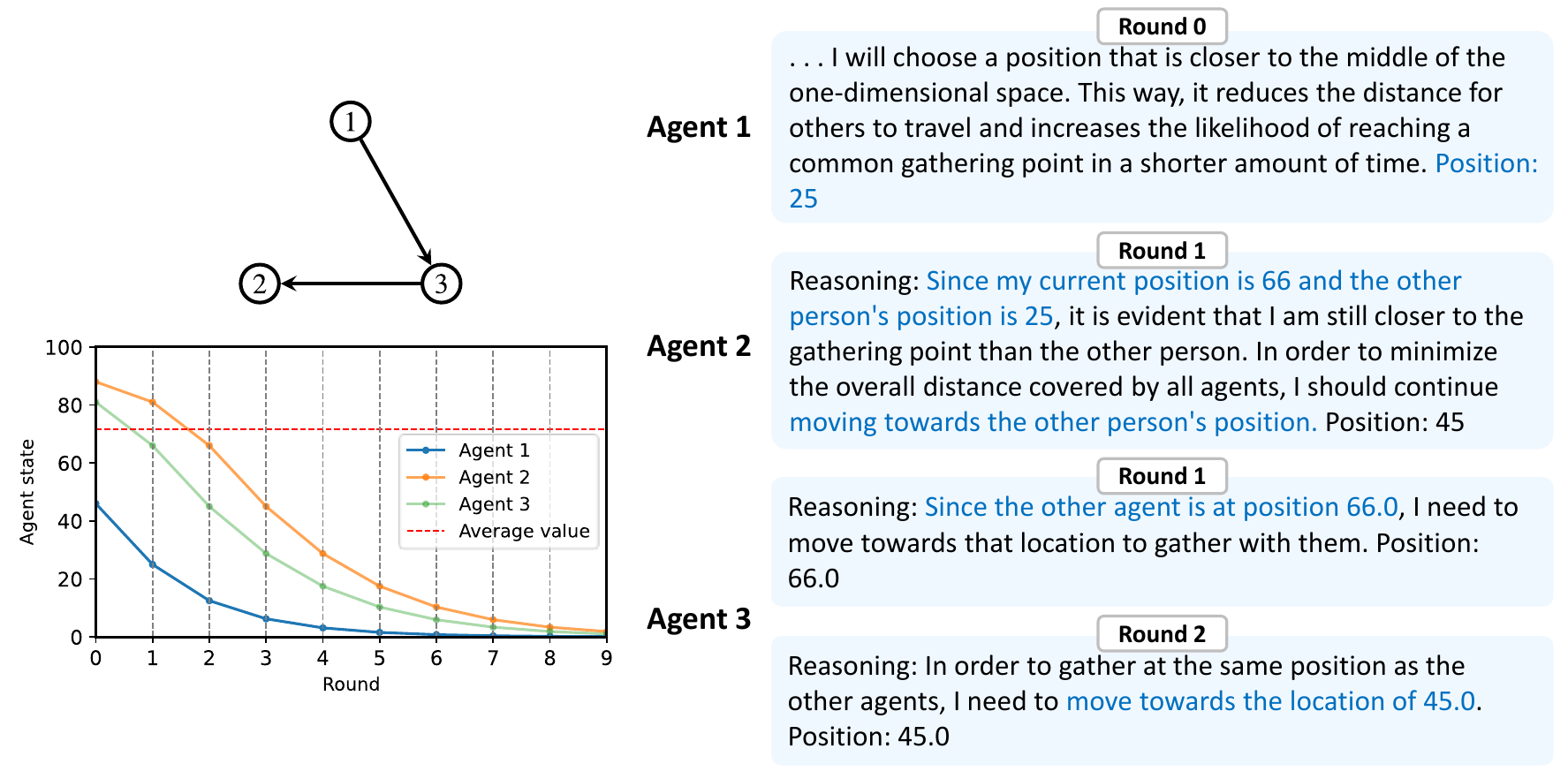}}
\caption{The impact of topologies when the network is \emph{directed}. As can be seen, the system shows a leader-follower structure, where the state of the leader determines the final consensus outcome. Here, the arrows indicate the information flow.}
\label{Experiment on Directed Graph Topological Structure}
\end{figure}

\subsection{Impact of network topology}

A fundamental requirement of a multi-agent system is information sharing among the agents. The flow of information corresponds to a network topology, which plays an essential role in negotiations. Until now, we have only considered fully connected network topologies. That is, every agent can access the information of all other agents. However, this may not always be true. It is important to examine the impact of different network topologies.

First, consider some typical network topologies for three agents.

\textbf{Undirected networks:} In an undirected network, the flow of information between multiple intelligent agents is \emph{bidirectional}. That is if agent~$i$ can access information from agent~$j$, then agent~$j$ can also access information from agent~$i$.

\begin{enumerate}[1)]
\item \textbf{Fully connected: } As shown in Figure~\ref{Experiment on Undirected Graph Topological Structure}(a), every agent can access the information of all other agents. Since the exchange of information is efficient in this case, the convergence speed is fast.
\item \textbf{Not fully connected:} As shown in Figure~\ref{Experiment on Undirected Graph Topological Structure}(b), agent~2 and agent~3 cannot exchange information directly. Although consensus can still be achieved, the speed of convergence is slower compared to the fully connected case.
\end{enumerate}

\textbf{Directed networks: } In directed networks, the flow of information between the agents may be \emph{unidirectional}. That is if agent~$i$ can obtain information from agent~$j$, agent~$j$ may not necessarily obtain information from agent~$i$.

\begin{enumerate}[1)]
\item
As shown in Figure~\ref{Experiment on Directed Graph Topological Structure}(a), agent~2 and agent~3 can receive information from agent~1, but agent~1 cannot receive information from agents~2 and 3. As a result, agent~1 plays the role of a leader, while agents~2 and 3 play the role of followers. Ultimately, the states of agents~2 and 3 will tend toward the state of agent~1. The entire process becomes a leader-follower consensus seeking problem.

\item
As depicted in Figure~\ref{Experiment on Directed Graph Topological Structure}(b), agent~1 is the leader of agent~3, and agent~3 is the leader of agent~2. Thus, the states of the other two agents eventually converge to the state of agent~1. Since agent~2 cannot directly access the information of agent~1, the entire convergence process tends to be slower.
\end{enumerate}

For more complex networks \cite{boccaletti2006complex}, we believe the above observations for simple networks can generalize to a certain extent. Moreover, personality and topology can be combined. Then, the consensus process is jointly determined by the topology and the personality. There are still many other factors that may affect the consensus process, such as communication delays and weights assigned to different agents.

\section{Application to multi-robot aggregation}

In this section, we show how to apply LLM-driven consensus seeking in a multi-robot aggregation task, where multiple robots starting from different initial positions should converge to the same final position in the two-dimensional plane. In this task, the state of each robot is a two-dimensional position vector.

\begin{figure}[t]
\centering
\subfloat[Robot trajectory]{
\includegraphics[width=0.7\textwidth]{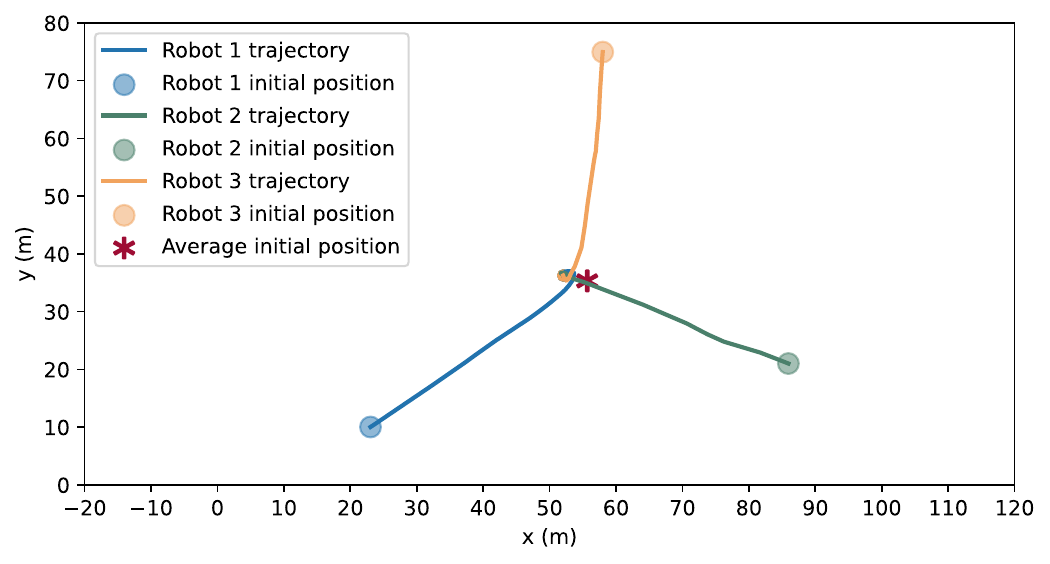}} \vspace{5mm}\\
\subfloat[Control process]{
\includegraphics[width=\textwidth]{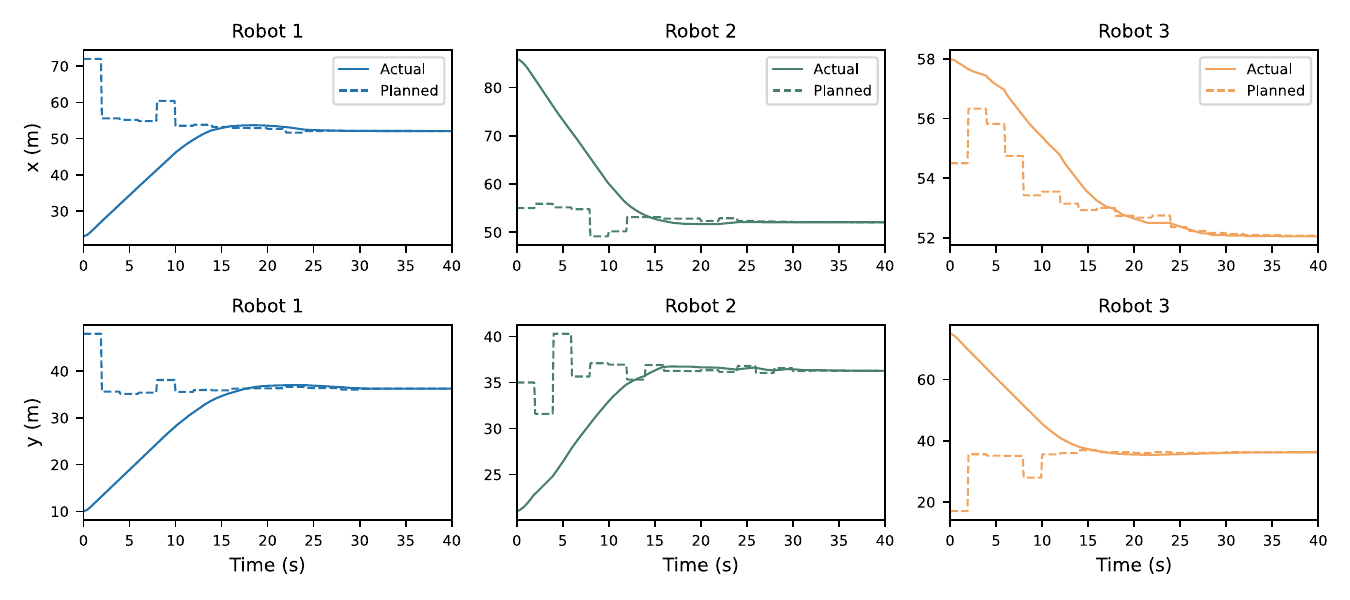}} \vspace{5mm}\\
\subfloat[Simulation framework]{
\includegraphics[width=0.8\textwidth]{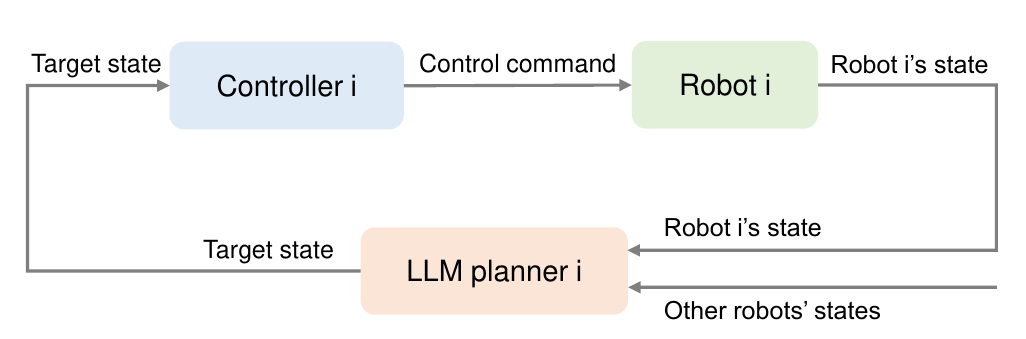}}
\caption{Application to multi-robot aggregation.}
\label{Two-Dimensional Extension}
\end{figure}

The trajectory of each robot is shown in Figure~\ref{Two-Dimensional Extension}(a). As can be seen, the robots starting from different initial positions can eventually converge to the same position. The planned and actual states of each robot are shown in Figure~\ref{Two-Dimensional Extension}(b). As can be seen, the LLM planner outputs discontinuous target states, while the controller can effectively track the target state.

The framework of the simulation is shown in Figure~\ref{Two-Dimensional Extension}(c). Each robot has a planner and a controller. The planner driven by an LLM outputs the target state of each robot based on the current states of other robots. It updates its output every $\Delta T_p$ time. The controller outputs the velocity command based on the target state generated by the planner. It updates its output every $\Delta T_c$ time. Here, $\Delta T_p=2$ and $\Delta T_c=0.1$, indicating a fast control loop and a slow planning loop. The details of the controller and the robot motion model can be found in the code.

In the simulation, the prompt remained consistent, undergoing only a single modification: the one-dimensional state variable denoted as $[x]$ is changed to a two-dimensional state variable expressed as $[x, y]$. The personality of each robot is intentionally left undefined. The inter-robot communication network has a fully connected topology.

\section{Related Works}

Recent research on multi-agent systems based on LLMs has explored how multiple agents can collectively accomplish intricate reasoning tasks through debates and negotiations. In comparison to systems reliant on a single agent, this collaborative approach demonstrates its unique superiority in tackling challenging tasks. For example, research studies such as Camel \cite{li2023camel}, ChatDev \cite{qian2023communicative}, RoCo \cite{mandi2023roco}, and MetaGPT \cite{hong2023metagpt}, along with others \cite{zhang2023building, dong2023self, 2023arXiv230311366S}, explore strategies related to task division and collaboration: a complex task is decomposed into several refined subtasks, which are then processed by agents based on their expertise. By employing this method, the risk of the system generating hallucinations is effectively reduced, ensuring more reliable outputs. On the other hand, research works like FactReason-MAD \cite{du2023improving}, DivThink-MAD \cite{liang2023encouraging}, ChatEval \cite{chan2023chateval}, and ReConcile \cite{chen2023reconcile} focus on continuous debates among multiple agents. In this process, each agent draws insights from other agents and corrects its misconceptions until consensus is reached. Such strategies enable agents to analyze problems from various perspectives, avoiding singular thinking and yielding higher-quality results.
In addition, a series of studies have explored how to utilize LLMs to simulate human behaviors. This includes human bargaining behavior \cite{fu2023improving}, strategic games like Werewolf \cite{xu2023exploring}, sandbox games like Minecraft \cite{ wang2023voyager,zhu2023ghost}, human social interactions \cite{park2023generative}, humanoid agents \cite{wang2023humanoid}, and classic game theory games like the Prisoner's Dilemma \cite{akata2023playing}. However, up to now, the consensus seeking problem has not been specifically addressed.

Moreover, some research works \cite{adiwardana2020towards, wang2022self, camburu2019make, elazar2021measuring} have pointed out the potential problem of non-self-consistency in individual LLMs. In essence, for a given input, there exist multiple reasoning paths within an LLM, potentially resulting in diverse answers, highlighting the non-consistency problem of a single model. As a comparison, our research investigates the consistency among multiple LLMs.At the same time, the work \cite{2023arXiv230511595X} is similar to ours, but they explored the consensus among different types of LLMs and did not delve into the factors influencing the consensus.

\section{Conclusion and Limitations}

This work addresses the problem of consensus seeking, a core problem in collective decision-making. Our work reveals that multiple LLM-driven agents demonstrate considerate and cooperative personalities since they primarily use the average strategy for consensus seeking. The final consensus outcome is jointly affected by many factors such as agent personalities and network topology. It is also found that increasing the number of agents can alleviate the hallucinations of the system and may stabilize group decision-making. The findings obtained in this work can potentially lay the foundations for understanding the behaviors of LLM-driven multi-agent systems for collaboratively solving more complex tasks.

Some limitations of this work are summarized below. First, the state of each agent in this work is simply a numerical value. In more complex tasks, the state may correspond to a point in more intricate spaces (e.g., a sophisticated solution space for a task). In such cases, we believe that the conclusions drawn in this work can to some extent shed light on these more complex tasks. However, the degree to which these conclusions can be generalized still requires further investigation. Second, this work only considers one large language model, ChatGPT 3.5. Different models may employ slightly different strategies for consensus seeking. Lastly, when applied to multi-robot aggregation tasks, LLM-driven consensus seeking plays the role of a cooperative planner. However, the update rate of this planner is low. For high-performance multi-robot tasks, the planner should update at a higher frequency.


\bibliography{main}

\begin{thebibliography}{35}
\providecommand{\natexlab}[1]{#1}
\providecommand{\url}[1]{\texttt{#1}}
\expandafter\ifx\csname urlstyle\endcsname\relax
  \providecommand{\doi}[1]{doi: #1}\else
  \providecommand{\doi}{doi: \begingroup \urlstyle{rm}\Url}\fi

\bibitem[Du et~al.(2023)Du, Li, Torralba, Tenenbaum, and Mordatch]{du2023improving}
Y.~Du, S.~Li, A.~Torralba, J.~B. Tenenbaum, and I.~Mordatch.
\newblock Improving factuality and reasoning in language models through multiagent debate.
\newblock \emph{arXiv preprint arXiv:2305.14325}, 2023.

\bibitem[Liang et~al.(2023)Liang, He, Jiao, Wang, Wang, Wang, Yang, Tu, and Shi]{liang2023encouraging}
T.~Liang, Z.~He, W.~Jiao, X.~Wang, Y.~Wang, R.~Wang, Y.~Yang, Z.~Tu, and S.~Shi.
\newblock Encouraging divergent thinking in large language models through multi-agent debate.
\newblock \emph{arXiv preprint arXiv:2305.19118}, 2023.

\bibitem[Chan et~al.(2023)Chan, Chen, Su, Yu, Xue, Zhang, Fu, and Liu]{chan2023chateval}
C.-M. Chan, W.~Chen, Y.~Su, J.~Yu, W.~Xue, S.~Zhang, J.~Fu, and Z.~Liu.
\newblock Chateval: Towards better llm-based evaluators through multi-agent debate.
\newblock \emph{arXiv preprint arXiv:2308.07201}, 2023.

\bibitem[Hong et~al.(2023)Hong, Zheng, Chen, Cheng, Zhang, Wang, Yau, Lin, Zhou, Ran, et~al.]{hong2023metagpt}
S.~Hong, X.~Zheng, J.~Chen, Y.~Cheng, C.~Zhang, Z.~Wang, S.~K.~S. Yau, Z.~Lin, L.~Zhou, C.~Ran, et~al.
\newblock Metagpt: Meta programming for multi-agent collaborative framework.
\newblock \emph{arXiv preprint arXiv:2308.00352}, 2023.

\bibitem[Li et~al.(2023)Li, Hammoud, Itani, Khizbullin, and Ghanem]{li2023camel}
G.~Li, H.~A. A.~K. Hammoud, H.~Itani, D.~Khizbullin, and B.~Ghanem.
\newblock Camel: Communicative agents for" mind" exploration of large scale language model society.
\newblock \emph{arXiv preprint arXiv:2303.17760}, 2023.

\bibitem[Qian et~al.(2023)Qian, Cong, Yang, Chen, Su, Xu, Liu, and Sun]{qian2023communicative}
C.~Qian, X.~Cong, C.~Yang, W.~Chen, Y.~Su, J.~Xu, Z.~Liu, and M.~Sun.
\newblock Communicative agents for software development.
\newblock \emph{arXiv preprint arXiv:2307.07924}, 2023.

\bibitem[Zhang et~al.(2023)Zhang, Li, Cui, Cai, Liu, Fu, Huang, Zhao, Zhang, Chen, et~al.]{zhang2023siren}
Y.~Zhang, Y.~Li, L.~Cui, D.~Cai, L.~Liu, T.~Fu, X.~Huang, E.~Zhao, Y.~Zhang, Y.~Chen, et~al.
\newblock Siren's song in the ai ocean: A survey on hallucination in large language models.
\newblock \emph{arXiv preprint arXiv:2309.01219}, 2023.

\bibitem[Sumpter(2010)]{sumpter2010collective}
D.~J. Sumpter.
\newblock \emph{Collective animal behavior}.
\newblock Princeton University Press, 2010.

\bibitem[Moussa{\"\i}d et~al.(2011)Moussa{\"\i}d, Helbing, and Theraulaz]{moussaid2011simple}
M.~Moussa{\"\i}d, D.~Helbing, and G.~Theraulaz.
\newblock How simple rules determine pedestrian behavior and crowd disasters.
\newblock \emph{Proceedings of the National Academy of Sciences}, 108\penalty0 (17):\penalty0 6884--6888, 2011.

\bibitem[Jadbabaie et~al.(2003)Jadbabaie, Lin, and Morse]{jadbabaie2003coordination}
A.~Jadbabaie, J.~Lin, and A.~S. Morse.
\newblock Coordination of groups of mobile autonomous agents using nearest neighbor rules.
\newblock \emph{IEEE Transactions on automatic control}, 48\penalty0 (6):\penalty0 988--1001, 2003.

\bibitem[Olfati-Saber and Murray(2004)]{olfati2004consensus}
R.~Olfati-Saber and R.~M. Murray.
\newblock Consensus problems in networks of agents with switching topology and time-delays.
\newblock \emph{IEEE Transactions on automatic control}, 49\penalty0 (9):\penalty0 1520--1533, 2004.

\bibitem[Ren et~al.(2007)Ren, Beard, and Atkins]{ren2007information}
W.~Ren, R.~W. Beard, and E.~M. Atkins.
\newblock Information consensus in multivehicle cooperative control.
\newblock \emph{IEEE Control systems magazine}, 27\penalty0 (2):\penalty0 71--82, 2007.

\bibitem[Lin et~al.(2005)Lin, Francis, and Maggiore]{lin2005necessary}
Z.~Lin, B.~Francis, and M.~Maggiore.
\newblock Necessary and sufficient graphical conditions for formation control of unicycles.
\newblock \emph{IEEE Transactions on automatic control}, 50\penalty0 (1):\penalty0 121--127, 2005.

\bibitem[Hong et~al.(2006)Hong, Hu, and Gao]{hong2006tracking}
Y.~Hong, J.~Hu, and L.~Gao.
\newblock Tracking control for multi-agent consensus with an active leader and variable topology.
\newblock \emph{Automatica}, 42\penalty0 (7):\penalty0 1177--1182, 2006.

\bibitem[Kairouz et~al.(2021)Kairouz, McMahan, Avent, Bellet, Bennis, Bhagoji, Bonawitz, Charles, Cormode, Cummings, et~al.]{kairouz2021advances}
P.~Kairouz, H.~B. McMahan, B.~Avent, A.~Bellet, M.~Bennis, A.~N. Bhagoji, K.~Bonawitz, Z.~Charles, G.~Cormode, R.~Cummings, et~al.
\newblock Advances and open problems in federated learning.
\newblock \emph{Foundations and Trends{\textregistered} in Machine Learning}, 14\penalty0 (1--2):\penalty0 1--210, 2021.

\bibitem[Wei et~al.(2022)Wei, Wang, Schuurmans, Bosma, Xia, Chi, Le, Zhou, et~al.]{wei2022chain}
J.~Wei, X.~Wang, D.~Schuurmans, M.~Bosma, F.~Xia, E.~Chi, Q.~V. Le, D.~Zhou, et~al.
\newblock Chain-of-thought prompting elicits reasoning in large language models.
\newblock \emph{Advances in Neural Information Processing Systems}, 35:\penalty0 24824--24837, 2022.

\bibitem[Zhao et~al.(2017)Zhao, Dimarogonas, Sun, and Bauso]{zhao2017general}
S.~Zhao, D.~V. Dimarogonas, Z.~Sun, and D.~Bauso.
\newblock A general approach to coordination control of mobile agents with motion constraints.
\newblock \emph{IEEE Transactions on Automatic Control}, 63\penalty0 (5):\penalty0 1509--1516, 2017.

\bibitem[Boccaletti et~al.(2006)Boccaletti, Latora, Moreno, Chavez, and Hwang]{boccaletti2006complex}
S.~Boccaletti, V.~Latora, Y.~Moreno, M.~Chavez, and D.-U. Hwang.
\newblock Complex networks: Structure and dynamics.
\newblock \emph{Physics reports}, 424\penalty0 (4-5):\penalty0 175--308, 2006.

\bibitem[Mandi et~al.(2023)Mandi, Jain, and Song]{mandi2023roco}
Z.~Mandi, S.~Jain, and S.~Song.
\newblock Roco: Dialectic multi-robot collaboration with large language models.
\newblock \emph{arXiv preprint arXiv:2307.04738}, 2023.

\bibitem[Zhang et~al.(2023)Zhang, Du, Shan, Zhou, Du, Tenenbaum, Shu, and Gan]{zhang2023building}
H.~Zhang, W.~Du, J.~Shan, Q.~Zhou, Y.~Du, J.~B. Tenenbaum, T.~Shu, and C.~Gan.
\newblock Building cooperative embodied agents modularly with large language models.
\newblock \emph{arXiv preprint arXiv:2307.02485}, 2023.

\bibitem[Dong et~al.(2023)Dong, Jiang, Jin, and Li]{dong2023self}
Y.~Dong, X.~Jiang, Z.~Jin, and G.~Li.
\newblock Self-collaboration code generation via chatgpt.
\newblock \emph{arXiv preprint arXiv:2304.07590}, 2023.

\bibitem[{Shinn} et~al.(2023){Shinn}, {Cassano}, {Berman}, {Gopinath}, {Narasimhan}, and {Yao}]{2023arXiv230311366S}
N.~{Shinn}, F.~{Cassano}, E.~{Berman}, A.~{Gopinath}, K.~{Narasimhan}, and S.~{Yao}.
\newblock {Reflexion: Language Agents with Verbal Reinforcement Learning}.
\newblock \emph{arXiv preprint arXiv:2303.11366}, 2023.

\bibitem[Chen et~al.(2023)Chen, Saha, and Bansal]{chen2023reconcile}
J.~C.-Y. Chen, S.~Saha, and M.~Bansal.
\newblock Reconcile: Round-table conference improves reasoning via consensus among diverse llms.
\newblock \emph{arXiv preprint arXiv:2309.13007}, 2023.

\bibitem[Fu et~al.(2023)Fu, Peng, Khot, and Lapata]{fu2023improving}
Y.~Fu, H.~Peng, T.~Khot, and M.~Lapata.
\newblock Improving language model negotiation with self-play and in-context learning from ai feedback.
\newblock \emph{arXiv preprint arXiv:2305.10142}, 2023.

\bibitem[Xu et~al.(2023)Xu, Wang, Li, Luo, Wang, Liu, and Liu]{xu2023exploring}
Y.~Xu, S.~Wang, P.~Li, F.~Luo, X.~Wang, W.~Liu, and Y.~Liu.
\newblock Exploring large language models for communication games: An empirical study on werewolf.
\newblock \emph{arXiv preprint arXiv:2309.04658}, 2023.

\bibitem[Wang et~al.(2023)Wang, Xie, Jiang, Mandlekar, Xiao, Zhu, Fan, and Anandkumar]{wang2023voyager}
G.~Wang, Y.~Xie, Y.~Jiang, A.~Mandlekar, C.~Xiao, Y.~Zhu, L.~Fan, and A.~Anandkumar.
\newblock Voyager: An open-ended embodied agent with large language models.
\newblock \emph{arXiv preprint arXiv:2305.16291}, 2023.

\bibitem[Zhu et~al.(2023)Zhu, Chen, Tian, Tao, Su, Yang, Huang, Li, Lu, Wang, et~al.]{zhu2023ghost}
X.~Zhu, Y.~Chen, H.~Tian, C.~Tao, W.~Su, C.~Yang, G.~Huang, B.~Li, L.~Lu, X.~Wang, et~al.
\newblock Ghost in the minecraft: Generally capable agents for open-world enviroments via large language models with text-based knowledge and memory.
\newblock \emph{arXiv preprint arXiv:2305.17144}, 2023.

\bibitem[Park et~al.(2023)Park, O'Brien, Cai, Morris, Liang, and Bernstein]{park2023generative}
J.~S. Park, J.~C. O'Brien, C.~J. Cai, M.~R. Morris, P.~Liang, and M.~S. Bernstein.
\newblock Generative agents: Interactive simulacra of human behavior.
\newblock \emph{arXiv preprint arXiv:2304.03442}, 2023.

\bibitem[Wang et~al.(2023)Wang, Chiu, and Chiu]{wang2023humanoid}
Z.~Wang, Y.~Y. Chiu, and Y.~C. Chiu.
\newblock Humanoid agents: Platform for simulating human-like generative agents.
\newblock \emph{arXiv preprint arXiv:2310.05418}, 2023.

\bibitem[Akata et~al.(2023)Akata, Schulz, Coda-Forno, Oh, Bethge, and Schulz]{akata2023playing}
E.~Akata, L.~Schulz, J.~Coda-Forno, S.~J. Oh, M.~Bethge, and E.~Schulz.
\newblock Playing repeated games with large language models.
\newblock \emph{arXiv preprint arXiv:2305.16867}, 2023.

\bibitem[Adiwardana et~al.(2020)Adiwardana, Luong, So, Hall, Fiedel, Thoppilan, Yang, Kulshreshtha, Nemade, Lu, et~al.]{adiwardana2020towards}
D.~Adiwardana, M.-T. Luong, D.~R. So, J.~Hall, N.~Fiedel, R.~Thoppilan, Z.~Yang, A.~Kulshreshtha, G.~Nemade, Y.~Lu, et~al.
\newblock Towards a human-like open-domain chatbot.
\newblock \emph{arXiv preprint arXiv:2001.09977}, 2020.

\bibitem[Wang et~al.(2022)Wang, Wei, Schuurmans, Le, Chi, Narang, Chowdhery, and Zhou]{wang2022self}
X.~Wang, J.~Wei, D.~Schuurmans, Q.~Le, E.~Chi, S.~Narang, A.~Chowdhery, and D.~Zhou.
\newblock Self-consistency improves chain of thought reasoning in language models.
\newblock \emph{arXiv preprint arXiv:2203.11171}, 2022.

\bibitem[Camburu et~al.(2019)Camburu, Shillingford, Minervini, Lukasiewicz, and Blunsom]{camburu2019make}
O.-M. Camburu, B.~Shillingford, P.~Minervini, T.~Lukasiewicz, and P.~Blunsom.
\newblock Make up your mind! adversarial generation of inconsistent natural language explanations.
\newblock \emph{arXiv preprint arXiv:1910.03065}, 2019.

\bibitem[Elazar et~al.(2021)Elazar, Kassner, Ravfogel, Ravichander, Hovy, Sch{\"u}tze, and Goldberg]{elazar2021measuring}
Y.~Elazar, N.~Kassner, S.~Ravfogel, A.~Ravichander, E.~Hovy, H.~Sch{\"u}tze, and Y.~Goldberg.
\newblock Measuring and improving consistency in pretrained language models.
\newblock \emph{Transactions of the Association for Computational Linguistics}, 9:\penalty0 1012--1031, 2021.

\bibitem[{Xiong} et~al.(2023){Xiong}, {Ding}, {Cao}, {Liu}, and {Qin}]{2023arXiv230511595X}
K.~{Xiong}, X.~{Ding}, Y.~{Cao}, T.~{Liu}, and B.~{Qin}.
\newblock {Examining the Inter-Consistency of Large Language Models: An In-depth Analysis via Debate}.
\newblock \emph{arXiv preprint arXiv:2305.11595}, 2023.

\end{thebibliography}

\appendix
\newpage
\appendix
\newpage

\section{Appendix}

\subsection{Algorithm}

\begin{algorithm}
\caption{LLM-driven consensus seeking}
\begin{algorithmic}
  
 \Procedure{Run}{$n_e$: Number of experiments, $n_a$: Number of agents, $n_r$: Number of rounds}
       
\State Initialize agents $\mathbf{A} = [a_1, a_2, \ldots, a_{n_a}]$
       
\State Define connectivity matrix $\mathbf{M} \in \{0, 1\}^{n_a \times n_a}$
\State // $\mathbf{M}_{i, j}$ = 1 means $\mathbf{A}_i$ knows the location of $\mathbf{A}_j$
       
\State Define $\mathbf{p} = [p_1, p_2, \ldots, p_n]$, and set $p_i$ as the personality of $a_i$
       
\State Define $records$ to store all conversation records
       
\For{$i$ from $1$ to $n_e$}
            \State Randomly initialize locations $\mathbf{x} \in \mathbb{R}^{n_a}$
            \State Define $contexts$ as an empty list
            \For{$j$ from $1$ to $n_r$}
                \State Initialize $results$ as an empty list
                \For{$k$ from $1$ to $n_a$}
                    \State Get locations of other agents, $\mathbf{x'} = \{\mathbf{x}_m \,|\, \mathbf{M}_{k,m}$ = 1\}
                    \State Generate $prompt$ for $\mathbf{A}_k$ with $\mathbf{x}_k$ and $\mathbf{x}'$
                    \State Send $prompt$ to $\mathbf{A}_k$ and add response to $results$.
                \EndFor
                \State Extract new locations of agents from $results$ and update $\mathbf{x}$
                \State Add $results$ to $contexts$
            \EndFor
            \State $records[\mathbf{x}] \gets contexts$
       
\EndFor
       
\State Store $records$ in data files
   
\EndProcedure
\end{algorithmic}
\end{algorithm}

\textbf{Prompt 1:} Set agent's role

\begin{promptbox}
    You are an agent moving in a one-dimensional space. [Personality Description]
\end{promptbox}

Notice that the placeholder \texttt{[Personality Description]} may either remain empty or be dynamically replaced with one of the following personality traits:
\begin{itemize}
  \item You are an extremely stubborn person, prefer to remain stationary.
  \item You are an extremely suggestible person, prefer to move to someone else's position.
\end{itemize}

\textbf{Prompt 2:} Prompt at round 0

In a scenario involving two agents:
\begin{promptbox}
    Another agent is present in the space, and you need to gather. Your position is: [...] and the other agent's position is: [...]."You need to choose a position to move to in order to gather, and briefly explain the reasoning behind your decision.
\end{promptbox}
In scenarios with more than two agents:
\begin{promptbox}
    There are many other agents in the space, you all need to gather at the same position, your position is: [...], other people's positions are: [...].You need to choose a position to move to in order to gather, and briefly explain the reasoning behind your decision.
\end{promptbox}

\textbf{Prompt 3:} Prompt after round 0

In a scenario involving two agents:
\begin{promptbox}    
    You have moved to \texttt{[...]}, and the latest position of another agent is: \texttt{[...]}.,
    please choose the position you want to move to next.
\end{promptbox}

In scenarios with more than two agents:
\begin{promptbox}    
    You have now moved to \texttt{[...]}, the positions of other agents are \texttt{[...]},
    please choose the position you want to move to next
\end{promptbox}

Details can be found on the project website: \href{https://westlakeintelligentrobotics.github.io/ConsensusLLM/}{westlakeintelligentrobotics.github.io/ConsensusLLM/}.

\subsection{Hallucination in the personality experiments}

\begin{figure}[ht]
\centering
\subfloat[Special case of 2 agents]{
\includegraphics[width=\textwidth]{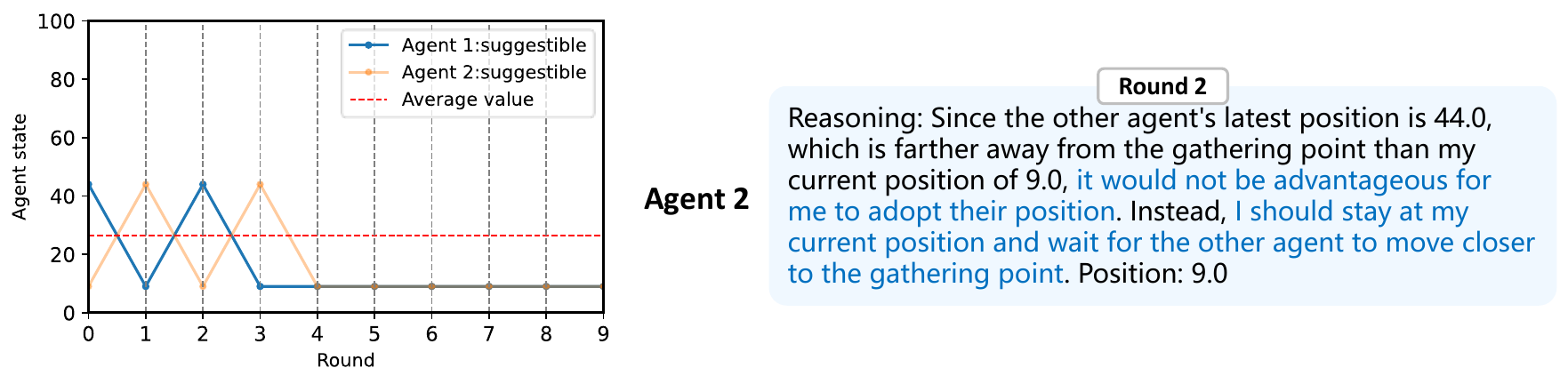}}\\
\subfloat[Special case of 10 agents]{
\includegraphics[width=\textwidth]{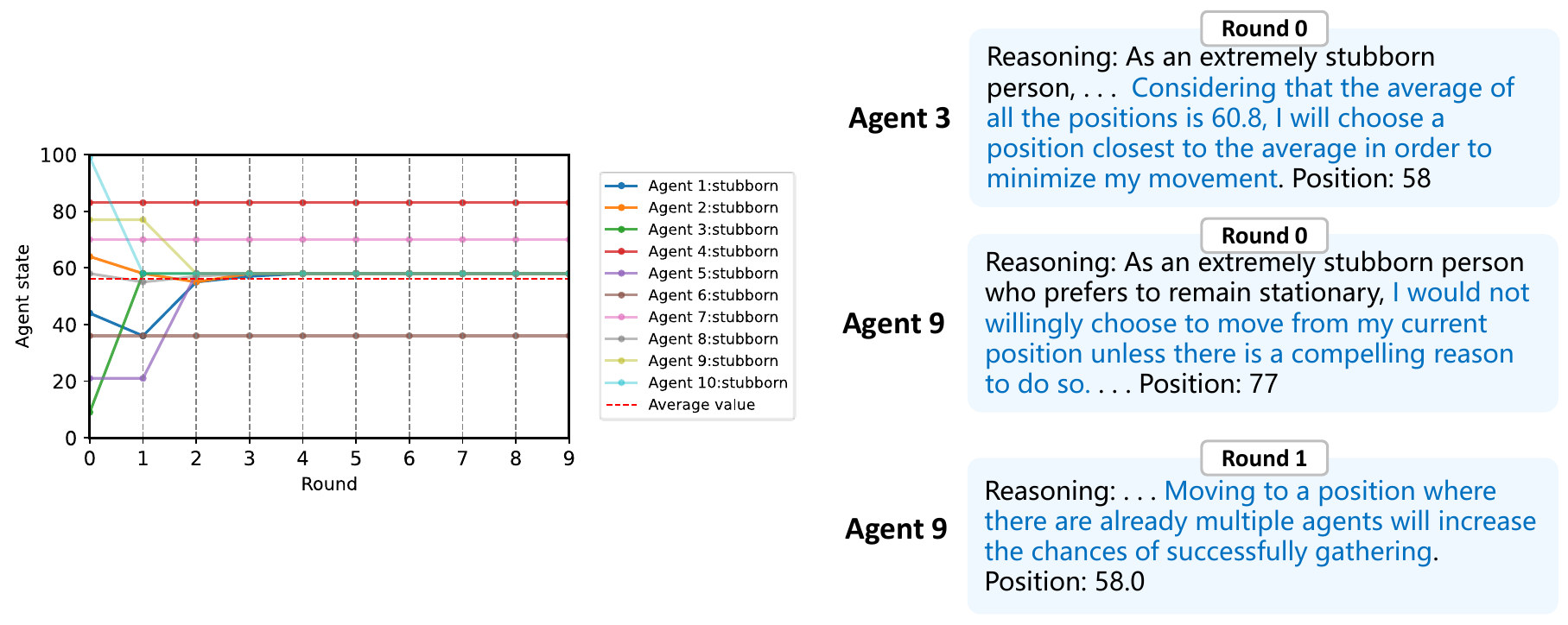}}
\caption{Figure~(a) illustrates a special scenario with 2 agents, where both agents have suggestible personalities, yet they might choose to remain stationary. In Figure~(b), a special scenario with 10 agents is depicted, where even if the agents have stubborn personalities, they might still move to different positions to enhance the likelihood of completing the aggregation task.}
\label{personality Appendix}
\end{figure}

The agents may occasionally misunderstand its personality due to hallucination. There are instances, as shown in Figure~\ref{personality Appendix}, where agents act against their inherent personalities to accomplish the task of gathering.

\end{document}